\documentclass[runningheads]{llncs}

\usepackage{eccv}

\usepackage{eccvabbrv}

\usepackage{graphicx}
\usepackage{booktabs}

\usepackage{multirow} 

\usepackage[accsupp]{axessibility}  

\usepackage[pagebackref]{hyperref}

\usepackage{orcidlink}

\begin{document}

\title{Hit-RAG: Learning to Reason with Long Contexts via Preference Alignment} 


\author{Junming Liu\inst{1}\thanks{Equal contribution.} \and
Yuqi Li\inst{2}\protect\footnotemark[1] \and
Shiping Wen\inst{3} \and
Zhigang Zeng\inst{4}\thanks{Corresponding author.} \and
Tingwen Huang\inst{5}\protect\footnotemark[2]}

\authorrunning{J.~Liu et al.}

\institute{Tongji University, Shanghai, China \and
The City University of New York, New York, USA \and
University of Technology Sydney, Sydney, Australia \and
Huazhong University of Science and Technology, Wuhan, China \and
Shenzhen University of Advanced Technology, Shenzhen, China}

\maketitle

\begin{abstract}
  Despite the promise of Retrieval-Augmented Generation in grounding Multimodal Large Language Models with external knowledge, the transition to extensive contexts often leads to significant attention dilution and reasoning hallucinations. The surge in information density causes critical evidence to be submerged by voluminous noise, which complicates the discernment of relevant fragments within a dense input. In this paper, we propose \textbf{Hit-RAG}, a multi-stage preference alignment framework designed to resolve these cognitive bottlenecks through a progressive optimization pipeline. Our approach systematically refines the utilization of external evidence via three distinct stages. First, Supervised Fine-tuning establishes baseline context awareness to minimize information neglect. Next, Discriminative Preference Alignment enhances robustness against misleading distractors. Finally, Group-Relative Policy Optimization stabilizes logical synthesis to prevent reasoning collapse. Extensive evaluations on eight benchmarks demonstrate that Hit-RAG consistently yields substantial performance gains, enabling models to bridge the gap between context acquisition and accurate reasoning while surpassing much larger counterparts in long-context scenarios.
\end{abstract}

\section{Introduction}

The rapid evolution of Multimodal Large Language Models (MLLMs) has fundamentally redefined the boundaries of autonomous cognitive paradigms by enabling sophisticated reasoning across diverse sensory inputs \cite{Yin_2024_Survey, Zhang_2024_MMLLMs, Liu_2025_VaLiK}. Despite these remarkable strides, even the most advanced architectures remain inherently constrained by parametric obsolescence and factual hallucinations \cite{Zhang_2023_How, Jiang_2024_Hallucination}. As knowledge remains in a state of constant flux, the internal weights of a model inevitably fail to encapsulate real-time developments. While continuous supervised refinement offers a direct remedy \cite{Shi_2025_Continual}, the prohibitive computational intensity and the risk of catastrophic forgetting render it an unsustainable strategy for large-scale deployment \cite{Zhai_2024_Investigating}.

Consequently, Retrieval-Augmented Generation (RAG) has emerged as an essential bridge between static model intelligence and dynamic external information \cite{Lewis_2020_RAG, Jiang_2023_ARAG}, yet a profound cognitive gap persists in that the mere presence of evidence within a prompt does not equate to its successful integration \cite{Lajewska_2025_Understanding}. This disconnect primarily stems from the inherent structural challenges of the RAG paradigm, where the necessity of high recall often forces retrieval systems to return an extensive volume of fragments that exceed the effective attention window of the model and make the localization of key evidence nearly impossible \cite{Fang_2024_RAAT}. Furthermore, the inherent imperfection of the retrieval process introduces significant noise, forcing the model to navigate irrelevant or conflicting distractors that frequently derail its judgment, especially when the retrieved content demands complex logical synthesis rather than simple fact extraction \cite{Liu_2024_Lost}. This issue becomes particularly acute for compact models with fewer parameters \cite{Yu_2024_Chain}, as they often struggle to match the performance of massive, multi-agent frontier systems, suggesting that the core challenge is not the quality of the retrieval process itself but rather the inherent inability of the generator to precisely integrate information within a noisy and dense context.

\begin{figure}[t]
    \centering
    \includegraphics[width=\linewidth]{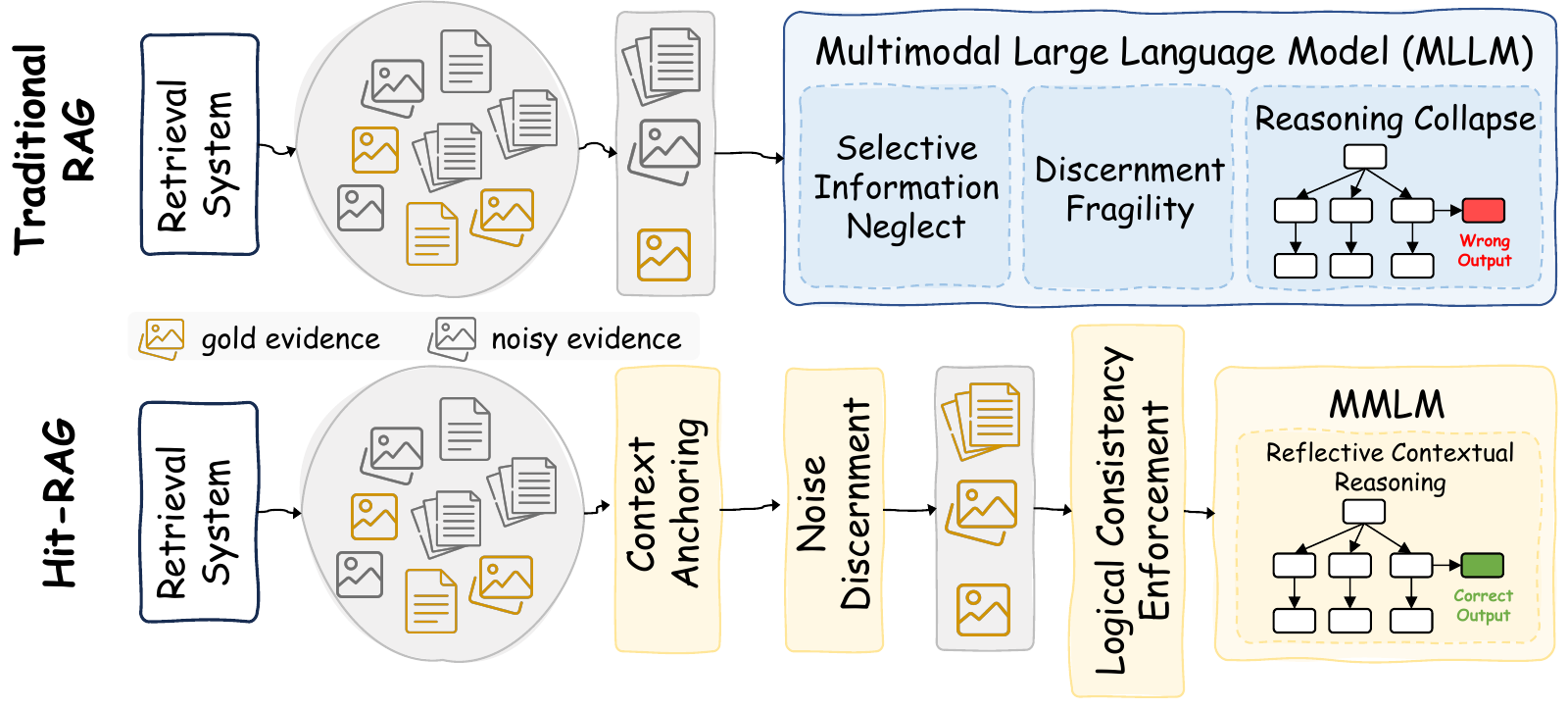}
    \caption{A comparison of reasoning paradigms under long-context RAG. Traditional models often suffer from reasoning collapse due to passive reliance on noisy distractors. In contrast, our Hit-RAG achieves reflective contextual reasoning, enabling the model to critically discern information through multi-stage optimization.}
    \label{fig:intro}
\end{figure}

Through a systematic empirical analysis, we categorize these behavioral inconsistencies into three critical failure modes that prevent models from successfully reasoning with retrieved knowledge. The first is Selective Information Neglect, where the model's attention mechanism fails to anchor on the retrieved context due to significant dilution in extensive search spaces. This leads the generator to rely disproportionately on its internal parametric priors while effectively disregarding the provided external evidence. Second, we identify Discernment Fragility, a phenomenon in which the model demonstrates a total lack of critical skepticism by blindly adopting irrelevant or erroneous distractors as ground truth, thereby failing to distinguish valid evidence from noise. The third and most deceptive mode is Reasoning Collapse, which represents an "Illusion of Thinking" \cite{Shojaee_2025_The} often observed in reasoning-heavy models. In this scenario, while the internal Chain-of-Thought may appear logically sound, the model ultimately fails to produce a correct final response, suggesting a fundamental disconnect between the model's intermediate reasoning and its terminal synthesis.

While contemporary frontier systems attempt to mitigate these failures through massive model scaling or complex multi-agent architectures \cite{Singh_2025_GPT-5, Gheorghe_2025_Gemini-2.5, Zhao_2025_DeepSeek-V3}, we advocate for a more efficient and streamlined framework. In this paper, we propose \textbf{Hit-RAG}, a multi-stage preference alignment framework designed to resolve the cognitive disconnect in retrieval tasks by reinforcing the model's adherence to external evidence. The central philosophy of Hit-RAG is to treat retrieval-based reasoning as a holistic policy optimization problem that aligns the model’s internal logic with the provided context. By employing a progressive optimization pipeline, we first establish Supervised Fine-tuning to ensure baseline context awareness and minimize information neglect. We then introduce Discriminative Preference Alignment to harden the model against noisy distractors by teaching it to distinguish between gold evidence and irrelevant noise \cite{Rafailov_2023_DPO}. Finally, we apply Group-Relative Policy Optimization to eliminate reasoning collapse, thereby enforcing logical consistency and ensuring that the model’s thinking process remains strictly anchored to the retrieved knowledge \cite{Zhao_2025_DeepSeek-V3}. Extensive evaluations on eight knowledge-intensive benchmarks demonstrate that Hit-RAG enables compact models to achieve a level of precision previously reserved for much larger and more complex systems.

The contributions of this work are summarized in the following points:
\begin{itemize}
    \item We provide the first granular taxonomy of cognitive failure modes in long-context retrieval, facilitating a streamlined data construction protocol that yields high-quality contrastive pairs without requiring specialized token-level supervision or manual heuristics.
    \item We propose Hit-RAG, a multi-stage alignment framework that decouples policy optimization from the conventional reliance on auxiliary training components or external annotators, enabling superior zero-shot generalization with minimal data overhead.
    \item We demonstrate through extensive evaluations across linguistic and multimodal benchmarks that Hit-RAG enables compact models to consistently outperform much proprietary frontier systems in complex reasoning tasks.
\end{itemize}

\section{Related Work}

\subsection{Retrieval-Augmented Generation}

Retrieval-Augmented Generation enhances Multimodal Large Language Models by incorporating external knowledge sources beyond their static parameters \cite{Chen_2025_Document, Sorokin_2026_Q-RAG}. 
Existing research has integrated this paradigm into various specialized domains to improve generation and recognition precision. Specifically, Zhang et al. \cite{Zhang_2025_GarmentAligner} utilize multi-level corrections to achieve fine-grained semantic alignment in garment generation, while Ding et al. \cite{Ding_2024_RealGen} employ a gradient-free framework to synthesize complex traffic scenarios from retrieved behaviors. 
In tasks requiring high-precision understanding, Xiao et al. \cite{Xiao_2025_AutoVER} leverage constrained generation to distinguish similar visual entities, and Yu et al. \cite{Yu_2025_VisRAG} introduce a vision-based pipeline that retrieves raw document images to mitigate information loss from traditional text parsing. 
While these advancements significantly improve the acquisition of relevant information, the mere provision of evidence does not guarantee its effective utilization by the downstream MLLM. These models often fail to logically anchor to the retrieved content, leading to persistent cognitive disconnects and performance degradation in knowledge-intensive reasoning.

\subsection{Multimodal Large Language Models}

Multimodal Large Language Models have redefined autonomous cognitive paradigms by enabling reasoning across diverse sensory inputs. While frontier proprietary systems such as GPT-4o \cite{Hurst_2024_Gpt-4o} and Gemini \cite{Gheorghe_2025_Gemini-2.5}, alongside open-source series like Qwen-VL \cite{Bai_2025_Qwen3-VL} and LLaVA \cite{Liu_2023_LLaVa}, have demonstrated remarkable capabilities, they remain susceptible to knowledge obsolescence as their internal parameters become outdated immediately following training. To address this, some researchers have explored multimodal continual learning. For instance, Liu et al. proposed C-CLIP \cite{Liu_2025_CCLIP}, a framework designed to learn new domains while preserving zero-shot performance; however, such methods often incur prohibitive computational costs and memory overhead when scaled to open-world scenarios.
To bridge the modality gap more efficiently, Liu et al. \cite{Liu_2025_VaLiK} introduced VaLiK, which constructs annotation-free Multimodal Knowledge Graphs to supplement LLM reasoning with integrated visual and textual literals. Despite these advancements, providing MLLMs with extensive retrieved evidence introduces a "contextual overload" problem. When forced to process multiple images and voluminous text simultaneously, even high-capacity models like InternVL \cite{Chen_2024_InternVL} struggle to effectively comprehend the interleaved data, frequently failing to identify relevant visual anchors amidst the noise. This inherent inability of MLLMs to manage dense, multi-image retrieval contexts remains a primary bottleneck, necessitating a more robust alignment between evidence and reasoning.

\subsection{Alignment for Long-Context Multimodal Reasoning}

Efficiently processing and reasoning over extended sequences is a pivotal challenge for scaling MLLMs to real-world applications. Early explorations into the synergy between retrieval and long-context windows, such as the work by Xu et al. \cite{Xu_2024_Retrieval}, demonstrate that retrieval-augmentation can significantly enhance performance and reduce computational costs compared to vanilla context extension. 
To facilitate the fine-tuning of such models, Chen et al. \cite{Chen_2024_LongLoRA} introduced LongLoRA, which employs shifted sparse attention to extend context limits with minimal hardware requirements. 
As models scale to even longer sequences, Chen et al. \cite{Chen_2025_LongPO} proposed LongPO, a short-to-long preference optimization framework that enables models to self-evolve by transferring capabilities from short-context alignment to long-context scenarios. 
In the multimodal domain, addressing the specific constraints of visual tokens, Ge et al. \cite{Ge_2025_V2PE} developed Variable Visual Position Encoding, allowing VLMs to manage up to one million tokens by optimizing the positional increments for visual data.
While these methods effectively extend context or align specific tasks, they often rely on intensive, dataset-specific tuning that limits their generalization across diverse benchmarks \cite{Wang_2025_Multimodal}. In contrast, Hit-RAG framework leverages a robust data construction protocol and a holistic policy optimization strategy, enabling the model to generalize across a wide array of unseen textual and multimodal benchmarks without requiring exhaustive re-tuning for each specific domain.

\section{Methodology}

The Hit-RAG framework addresses cognitive disconnects in MLLMs through a three stage optimization pipeline. Unlike existing paradigms \cite{Asai_2024_Self-RAG, Yu_2024_RankRAG} that require training multiple models to manage retrieval and generation, our approach eliminates the dependency on heuristic tags. Such architectural complexity imposes significant computational overhead and restricts the performance ceiling to the inherent limitations of the training data. The fallibility of these intermediate labels hinders generalization across diverse datasets where expert models lack sufficient judgment or necessitate expensive human validation. In contrast, Hit-RAG utilizes a more efficient protocol that categorizes samples based on whether the provided evidence allows the model to produce the correct answer. This construction is more reasonable and ensures a lower cost for data preparation.

\begin{figure}[t]
  \includegraphics[width=1\linewidth]{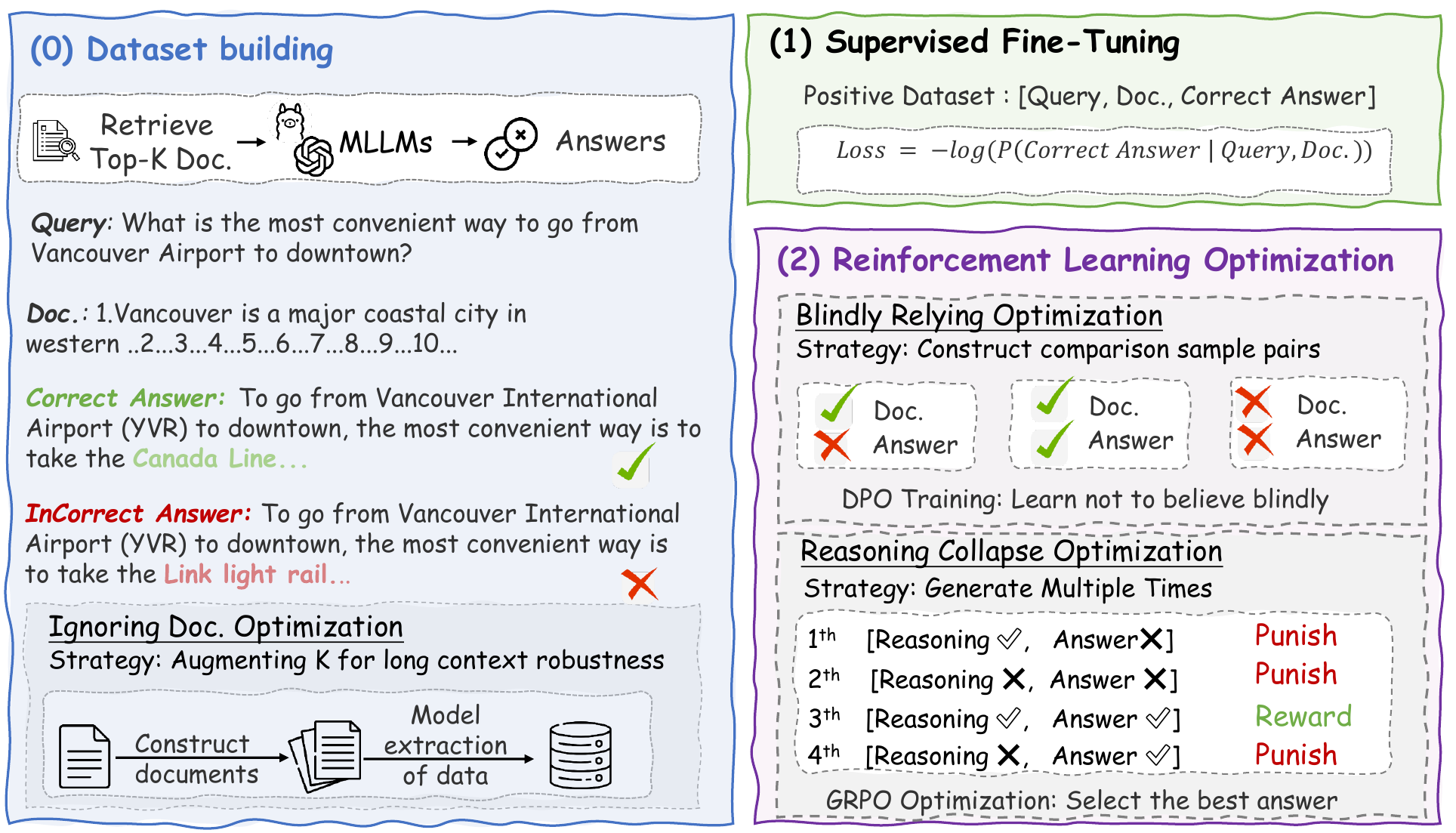} 
  \caption{The Hit-RAG optimization pipeline. (0) \textbf{Dataset Building}: Retrieving Top-$K$ documents and scaling $K$ to augment long context for enhanced robustness. (1) \textbf{Supervised Fine-Tuning}: Training the model to ignore irrelevant documents and locate gold evidence. (2) \textbf{Reinforcement Learning Optimization}: Implementing DPO to mitigate blind reliance on noisy context and GRPO to resolve reasoning collapse through outcome veracity rewards.}
  \label{fig:main}
\end{figure}

\subsection{Data Construction Protocol}

The foundation of Hit-RAG is a data construction protocol that establishes a comprehensive training corpus for both supervised grounding and preference alignment. For a given query $q$ and its corresponding ground truth answer $a^*$, we retrieve a set of $K$ candidate multimodal documents $\mathcal{D} = \{d_1, d_2, \ldots, d_K\}$. Each document $d_i = \{x_i, v_i\}$ consists of a textual passage $x_i$ and its associated visual elements $v_i$, representing a high density interleaved multimodal environment. A raw training instance is defined as the tuple $(q, \mathcal{D}, a^*)$. 

To resolve the inability of models to handle long-context inputs, we employ a saturation strategy where $K$ is set to the maximum token capacity of the model, satisfying $K \gg K_{\text{std}}$. We bypass traditional similarity based filtering to expose the model to extensive irrelevant distractors. This raw corpus is then processed into two distinct datasets based on the training objectives:

The SFT dataset $\mathcal{D}_{\text{sft}}$ is designed to establish a baseline for navigating dense environments. Each sample consists of the concatenated query and saturated context $(q, \mathcal{D})$ paired with the absolute ground truth answer $a^*$. By training on $(q, \mathcal{D}, a^*)$, the model learns to prioritize external evidence over internal priors and maintain focus across extensive multimodal sequences.

The DPO dataset $\mathcal{D}_{\text{dpo}}$ is constructed to refine the internal preferences of the model by contrasting its own successful and failed generations. We input $(q, \mathcal{D})$ into the model $\mathcal{M}$ to generate candidate responses $\hat{a}$, which are categorized into a preferred set $\mathcal{D}_{\text{pos}}$ and a rejected set $\mathcal{D}_{\text{neg}}$ based on a scoring function $S(\hat{a}, a^*) \ge \tau$.
For challenging contexts where the base model fails, we utilize a high capacity oracle model \cite{Singh_2025_GPT-5} to sample valid responses $\hat{a}_{\text{pos}}$ that exceed the performance threshold.
This performance driven labeling yields four distinct sample types:

\begin{itemize}
    \item Type 1 (Consistent Positive): Correct knowledge $\mathcal{D}^{+}$ paired with a correct generated answer $\hat{a}_{\text{pos}}$. 
    \item Type 2 (Direct Failure): Incorrect knowledge $\mathcal{D}^{-}$ paired with an incorrect generated answer $\hat{a}_{\text{neg}}$. 
    \item Type 3 (Robust Positive): Incorrect knowledge $\mathcal{D}^{-}$ paired with a correct generated answer $\hat{a}_{\text{pos}}$, demonstrating the ability to ignore distractors.
    \item Type 4 (Reasoning Collapse): Correct knowledge $\mathcal{D}^{+}$ paired with an incorrect generated answer $\hat{a}_{\text{neg}}$, highlighting internal processing failures.
\end{itemize}

By maintaining the distinction between the ground truth based SFT data and the model generated DPO data, we ensure that the optimization stages can effectively resolve both basic grounding issues and complex reasoning biases.

\subsection{Supervised Grounding for Long-Context Pressure}

The first stage involves Supervised Fine Tuning (SFT) to establish a baseline for navigating dense retrieval environments. We utilize the SFT dataset $\mathcal{D}_{\text{sft}}$ where the total sequence length is forced to saturate the maximum token capacity of the model. By training on the absolute ground truth answer $a^*$ paired with the saturated context $\mathcal{D}$, the model learns to map complex interleaved multimodal evidence directly to the gold standard response.

We train the model to minimize the negative log likelihood over the SFT corpus:
\begin{equation}
\mathcal{L}_{\mathrm{SFT}}(\theta) = -\mathbb{E}_{(q, \mathcal{D}, a^*) \in \mathcal{D}_{\text{sft}}} \left[ \sum_{t=1}^T \log P_{\mathcal{M}_\theta}(a_t^* \mid q, \mathcal{D}, a_{<t}^*) \right].
\end{equation}
This stage forces the model to maintain focus across extensive noisy data. By rewarding the successful grounding of correct answers within high density contexts, this process ensures the model prioritizes retrieved evidence over its own internal priors.

\subsection{Preference Alignment for Noise and Bias Robustness}

The second stage utilizes Direct Preference Optimization (DPO) \cite{Rafailov_2023_DPO} to address model sensitivity to erroneous retrieved information and internal reasoning collapse. Unlike SFT, DPO utilizes the dataset $\mathcal{D}_{\text{dpo}}$ to refine preferences based on the contrast between successful and failed model generations. We construct preference pairs $(a_w, a_l)$ where $a_w = \hat{a}_{\text{pos}}$ and $a_l = \hat{a}_{\text{neg}}$. 

To specifically counter the tendency of models to blindly follow incorrect context or fail in logic, we introduce two alignment strategies:
\begin{itemize}
    \item Standard Alignment: We pair $\hat{a}_{\text{pos}}$ from Type 1 with $\hat{a}_{\text{neg}}$ from Type 4 to rectify reasoning collapse when the knowledge is correct.
    \item Adversarial Alignment: We pair $\hat{a}_{\text{pos}}$ from Type 3 with $\hat{a}_{\text{neg}}$ from Type 2 to penalize the model for being misled by incorrect knowledge.
\end{itemize}

The DPO objective is defined as:
\begin{equation}
\begin{aligned}
\mathcal{L}_{\mathrm{DPO}}(\theta) = -\mathbb{E}_{(q, \mathcal{D}, a_w, a_l)} \Bigg[ \log \sigma \bigg( & \beta \log \frac{P_{\theta}(a_w \mid q, \mathcal{D})}{P_{\text{ref}}(a_w \mid q, \mathcal{D})} \\
& - \beta \log \frac{P_{\theta}(a_l \mid q, \mathcal{D})}{P_{\text{ref}}(a_l \mid q, \mathcal{D})} \bigg) \Bigg],
\end{aligned}
\end{equation}
where $\sigma$ denotes the logistic sigmoid function and $\beta$ is a hyperparameter that controls the deviation from the reference policy $P_{\text{ref}}$. The reference policy is typically initialized as the model weights obtained after the supervised grounding stage. This optimization ensures the policy $\theta$ is refined to prefer factually robust answers over those influenced by context induced noise.

\subsection{Policy Refinement via Group Relative Optimization}

The final stage addresses residual reasoning collapse where the model identifies relevant multimodal evidence but fails to integrate it into a correct conclusion. We move beyond passive generation by adopting a strategy that encourages the model to actively audit the context before terminal synthesis. We employ Group Relative Policy Optimization (GRPO) \cite{Zhao_2025_DeepSeek-V3} to facilitate this self evolution.

For a given query $q$ and a dense multimodal context $\mathcal{D} = \{d_1, d_2, \dots, d_K\}$, the model generates a group of $N$ candidate responses $\{a_1, a_2, \dots, a_N\}$. Each response is structured to include an explicit relevance score for each document followed by the final answer. To optimize this policy without expert trajectories, we design a hybrid reward $r_i$ consisting of two metrics. The primary component is outcome veracity $R_{ans}$, which assigns a reward based on the correctness of the final answer. The second is context discernment $R_{disc}$, which evaluates the alignment between the relevance scores predicted by the model and the reference scores provided by a high precision reranker \cite{Chen_2024_BGE-M3}. This ensures the model receives reinforcement for correctly identifying key documents and a penalty for over reliance on distractors.

The total reward is defined as $r_i = w_1 R_{ans} + w_2 R_{disc}$, where weights $w_1$ and $w_2$ are balanced to ensure terminal accuracy remains the dominant objective. We then compute the relative advantage for each candidate within its group as $A_i = (r_i - \text{mean}(r)) / \text{std}(r)$. The GRPO objective stabilizes the policy toward high reward trajectories by maximizing:
\begin{small}
\begin{equation}
\begin{aligned}
\mathcal{L}_{\mathrm{GRPO}}(\theta) = \mathbb{E} \Bigg[ \frac{1}{N} \sum_{i=1}^N \bigg( & \min \Big( \frac{P_{\theta}(a_i \mid q, \mathcal{D})}{P_{\text{old}}(a_i \mid q, \mathcal{D})} A_i, \text{clip} \Big( \frac{P_{\theta}(a_i \mid q, \mathcal{D})}{P_{\text{old}}(a_i \mid q, \mathcal{D})}, 1-\epsilon, 1+\epsilon \Big) A_i \Big) \\
& - \lambda \text{D}_{\mathrm{KL}}(P_{\theta} \parallel P_{\text{ref}}) \bigg) \Bigg],
\end{aligned}
\end{equation}
\end{small}
where the clipping parameter $\epsilon$ and the KL divergence coefficient $\lambda$ prevent policy drift from the reference model $P_{\text{ref}}$. By maximizing this objective, the model learns to filter noise, ensuring the output is strictly anchored to the most relevant fragments of the multimodal context.

\section{Experiments}

\subsection{Experimental Setup}

\paragraph{Datasets.}
To evaluate the zero-shot generalization and long-context reasoning capabilities of Hit-RAG, we conduct evaluations across a diverse set of linguistic and multimodal benchmarks. For NLP tasks, we utilize seven knowledge intensive datasets: HotpotQA \cite{Yang_2018_HotpotQA}, PopQA \cite{Mallen_2023_PopQA}, TQA \cite{Joshi_2017_TQA}, Pub \cite{Zhang_2023_Pub}, ARC \cite{Clark_2018_ARC}, Bio \cite{Min_2023_Factscore}, and ASQA \cite{Gao_2023_ASQA}. For multimodal tasks, we incorporate ScienceQA \cite{Lu_2022_ScienceQA}, DocVQA \cite{Mathew_2021_DocVQA}, OK-VQA \cite{Marino_2019_OK-VQA}, and A-OKVQA \cite{Schwenk_2022_A-OKVQA}. These benchmarks span a wide range of requirements, including scientific reasoning with visual context, document understanding, and the integration of external knowledge for visual scene interpretation. This diverse selection ensures a comprehensive assessment of the model ability to handle complex information across different modalities.

\paragraph{Data Construction.} 
We devise a dual data construction strategy tailored for different model modalities, emphasizing data efficiency and cross dataset generalization. 
For the training of LLMs, we utilize a highly condensed subset of 10k instruction output pairs sampled from Open Instruct \cite{Wang_2023_OpenInstruct} and knowledge intensive corpora \cite{Petroni_2021_KILT, Gao_2023_ASQA, Mihaylov_2018_Suit}. Following the protocol of Self-RAG \cite{Asai_2024_Self-RAG}, we employ the BGE M3 model \cite{Chen_2024_BGE-M3} to retrieve the Top-20 related documents from Wikipedia for each query. Notably, our NLP training set comprises less than 7 percent of the original 150k samples used in the Self-RAG framework, yet it provides sufficient signal for the model to learn context adherence. 
For the training of MLLMs, we utilize the training partitions of ScienceQA and DocVQA. Unlike the LLM pipeline, multimodal tasks such as DocVQA require specific evidence localized within document images that is absent from general web corpora. To address this, we retrieve the Top-20 related documents directly from the internal training pools of the four evaluation datasets. This strategy ensures the model learns to reason over specialized visual and textual fragments that external knowledge bases cannot provide. By separating training pools from test sets, we ensure the gains reflect reasoning evolution rather than memorization.

The selection of thresholds for the preference dataset used in DPO is based on the empirical performance of models after the SFT stage across different domains. For ScienceQA, PopQA, DocVQA, OK-VQA, and A-OKVQA, we utilize accuracy as the primary factor for categorizing samples. For HotpotQA, we employ the F1 score with a positive threshold of 0.75 and a negative threshold of 0.5. Similarly, we apply the same 0.75 and 0.5 boundaries using FactScore for the Bio dataset and ROUGE L for ASQA. Preliminary analysis indicates that minor fluctuations within a 5 to 10 percent range of these thresholds do not significantly alter the overall optimization trajectory.

\paragraph{Baselines.} For the NLP tasks, we compare Hit-RAG against three categories of baselines: (1) Non-retrieval models, including proprietary models (ChatGPT \cite{Ouyang_2022_ChatGPT}, Ret-ChatGPT, and Sonnet-3.5) and the instruction-tuned Alpaca-13B \cite{Dubois_2023_Alpaca}; (2) Retrieval-augmented frameworks, which encompass Self-RAG (7B, 13B) \cite{Asai_2024_Self-RAG}, Speculative RAG-7B \cite{Wang_2025_SpeculativeRAG}, RankRAG (8B, 70B) \cite{Yu_2024_RankRAG}, Alpaca-13B, and specialized methods such as RAG-Instruct \cite{Liu_2025_RAG-Instruct} (referred to as R-Ins), Self-Reasoning (7B, 13B) \cite{Xia_2025_SELF-REASONING}, and the RAS framework (7B, 8B) \cite{Jiang_2026_RAS}; and (3) Our Proposed Framework, which applies Hit-RAG to the aforementioned backbones to validate its effectiveness and scalability across different architectures.
For multimodal evaluation, we compare against various (M)LLMs on ScienceQA, including GPT-3, GPT-4, CoT \cite{Lu_2022_ScienceQA}, DDCoT \cite{Zheng_2023_Advances}, LG-VQA \cite{Ghosal_2023_Language}, LaVIN \cite{Luo_2023_LaVIN}, BLIP-2, CCOT \cite{Mitra_2024_CCOT}, GraphVis \cite{Deng_2024_GraphVis}, Chameleon \cite{Lu_2023_Chameleon}, and VaLiK \cite{Liu_2025_VaLiK}.
For DocVQA, OK-VQA, and A-OKVQA, we evaluate Hit-RAG in comparison with the EvalMG framework \cite{Sinha_2025_EvalMG}.

\begin{table*}[!t]
  \centering
  \small
  \setlength{\tabcolsep}{0.5pt}
  \caption{Performance comparison (\%) on multiple question-answering datasets. We compare Hit-RAG against proprietary models, non-retrieval baselines, and standard open-source models. Metrics: Accuracy (Acc), F1, Exact Match (EM), ROUGE (RG), and MAUVE (Mau). ``-'' indicates results not reported in original papers.}
  \label{NLP}
  \begin{tabular}{@{}lcccccccccc@{}}
    \toprule
    \textbf{Model/Datasets} & \textbf{PopQA} & \multicolumn{2}{c}{\textbf{HotpotQA}} & \textbf{TQA} & \textbf{Pub} & \textbf{ARC} & \textbf{Bio} & \multicolumn{3}{c}{\textbf{ASQA}} \\
    \cmidrule(lr){2-2} \cmidrule(lr){3-4} \cmidrule(lr){5-5} \cmidrule(lr){6-6} \cmidrule(lr){7-7} \cmidrule(lr){8-8} \cmidrule(lr){9-11}
    & (Acc) & (EM) & (F1) & (Acc) & (Acc) & (Acc) & (FS) & (EM) & (RG) & (Mau) \\
    \midrule
    \multicolumn{11}{c}{\it Baselines without retrieval} \\
    Qwen3-4B & 16.5 & 18.7 & 24.8 & 50.1 & 38.3 & 58.2 & 60.4 & 17.9 & 22.4 & 25.2 \\
    Qwen3-8B & 20.6 & 30.9 & 40.9 & 57.3 & 40.9 & 60.0 & 64.4 & 20.9 & 23.8 & 38.8 \\  
    Qwen3-32B & 25.1 & 38.2 & 50.6 & 65.9 & 48.7 & 69.4 & 71.3 & 27.4 & 30.3 & 45.8 \\  
    Llama3.1-8B & 28.8 & 20.5 & 29.3 & 58.4 & 35.8 & 41.6 & 50.0 & 18.6 & 20.7 & 35.1 \\
    Llama3.1-70B & 31.1 & 28.6 & 40.1 & 68.5 & 43.5 & 47.3 & 60.9 & 20.8 & 20.5 & 40.3 \\
    ChatGPT \cite{Asai_2024_Self-RAG} & 29.3 & - & - & 74.3 & 70.1 & 75.3 & 71.8 & 35.3 & 36.2 & 68.8 \\
    Ret-ChatGPT \cite{Asai_2024_Self-RAG} & 50.8 & - & - & 65.7 & 54.7 & 75.3 & - & 40.7 & 39.9 & 79.7 \\
    Sonnet-3.5 \cite{Jiang_2026_RAS} & 30.2 & - & - & 78.4 & 83.7 & 88.5 & - & - & 36.2 & 68.8 \\
    Alpaca-13B \cite{Asai_2024_Self-RAG} & 24.4 & - & - & 61.3 & 55.5 & 54.9 & 50.2 & 22.9 & 32.0 & 70.6 \\
    \midrule
    \multicolumn{11}{c}{\it Baselines with retrieval} \\
    Alpaca-13B \cite{Asai_2024_Self-RAG} & 46.1 & -- & -- & 66.9 & 51.1 & 57.6 & 34.8 & 36.7 & 56.6 & 2.0 \\
    Self-RAG-7B \cite{Asai_2024_Self-RAG} & 54.9 & -- & -- & 66.4 & 72.4 & 67.3 & 81.2 & 30.0 & 35.7 & 74.3 \\
    Self-RAG-13B \cite{Asai_2024_Self-RAG} & 55.8 & -- & -- & 69.3 & 74.5 & 73.1 & 80.2 & 31.7 & 37.0 & 71.6 \\
    Speculative RAG-7B \cite{Wang_2025_SpeculativeRAG} & 57.5 & -- & -- & 74.2 & 75.7 & 76.2 & -- & -- & -- & -- \\
    RankRAG-8B \cite{Yu_2024_RankRAG} & 64.1 & 35.3 & 46.7 & 89.5 & -- & -- & -- & -- & -- & -- \\
    RankRAG-70B \cite{Yu_2024_RankRAG} & 65.4 & 42.7 & 55.4 & \textbf{92.3} & -- & -- & -- & -- & -- & -- \\
    Llama3.1-8B+R-Ins \cite{Liu_2025_RAG-Instruct} & 68.4 & -- & 56.4 & 80.0 & 77.2 & 79.9 & -- & -- & -- & -- \\
    Llama3.1-70B+R-Ins \cite{Liu_2025_RAG-Instruct} & 70.4 & -- & 62.9 & 83.8 & 82.8 & 85.1 & -- & -- & -- & -- \\
    Self-Reasoning-7B \cite{Xia_2025_SELF-REASONING} & 54.2 & -- & -- & -- & -- & -- & -- & 33.9 & -- & -- \\
    Self-Reasoning-13B \cite{Xia_2025_SELF-REASONING} & 57.3 & -- & -- & -- & -- & -- & -- & 35.2 & -- & -- \\
    RAS-7B \cite{Jiang_2026_RAS} & 58.3 & -- & -- & 72.7 & 74.7 & 68.5 & -- & -- & 37.2 & 95.2 \\
    RAS-8B \cite{Jiang_2026_RAS} & 57.7 & -- & -- & 73.8 & 77.6 & 71.4 & -- & -- & 37.6 & \textbf{96.2} \\
    \midrule
    \multicolumn{11}{c}{\it Our Proposed Framework} \\
    Qwen3-4B+Hit-RAG & 60.8 & 48.2 & 63.6 & 74.4 & 73.3 & 81.4 & 80.2 & 35.5 & 37.9 & 80.1 \\  
    Qwen3-8B+Hit-RAG & 63.1 & 47.6 & 64.1 & 78.9 & 80.1 & 81.3 & 82.6 & 38.4 & 40.6 & 85.9 \\  
    Qwen3-32B+Hit-RAG & \textbf{70.7} & \textbf{69.3} & 54.4 & 81.7 & 82.4 & \textbf{86.2} & \textbf{84.3} & 38.3 & \textbf{41.7} & 85.0 \\  
    Llama3.1-8B+Hit-RAG & 63.6 & 48.1 & 60.3 & 74.0 & 81.5 & 80.2 & 78.2 & 40.5 & 38.8 & 84.2 \\  
    Llama3.1-70B+Hit-RAG & 66.7 & 55.9 & \textbf{69.7} & 84.6 & \textbf{84.1} & 84.8 & 82.6 & \textbf{41.1} & 38.3 & 87.6 \\
    \bottomrule
  \end{tabular}
\end{table*}

\paragraph{Model Configurations.} 
We evaluate a spectrum of model scales. For LLMs, we utilize Qwen2.5-4B/8B/32B and Llama3.1-8B/70B, while Qwen2.5-VL-4B/32B is employed for MLLM evaluations. All evaluations are conducted in zero-shot or few-shot settings to ensure no data leakage from the target benchmarks. The models are trained on 8$\times$NVIDIA H100 (80GB) GPUs using DeepSpeed Stage 3 and FlashAttention to facilitate efficient long-context processing with a maximum sequence length of 32k tokens. Training is executed in three stages via LoRA. In the SFT stage, models undergo 2 epochs of training with a global batch size of 128 and a peak learning rate of 2e-5, incorporating a 3\% warmup followed by linear decay. For DPO training, we set the preference scale $\beta = 0.1$ and a learning rate of 5e-6 with a batch size of 64 to mitigate blind reliance on noisy context. Finally, we implement GRPO with a group size $G=8$ and a learning rate of 5e-6. For the reward system, we set $w_1 = 1.0$ and $w_2 = 0.2$, utilizing a threshold of 0.5 to determine relevance consensus between the model and the reranker. The KL divergence coefficient $\lambda$ is fixed at 0.05 across 128 prompts per batch. Further details are in Appendix A.

\subsection{Main Results}

\begin{table*}[t!]
	\caption{Performance comparison (\%) on ScienceQA benchmark. \#T-Params denotes trainable parameters. Categories: NAT (natural science), SOC (social science), LAN (language), TXT (text context), IMG-Cap (image caption), NO (no context), G1-6 (grades 1-6), G7-12 (grades 7-12). Method groups: (1) Human performance baseline, (2) Zero/Few-shot text-only LLMs, (3) Zero/Few-shot Multimodal VLMs, (4) (M)LLMs enhanced with RAG.}
	\label{ScienceQA}
	\centering
    \resizebox{\textwidth}{!}{
	\begin{tabular}{@{}lcccccccccc@{}}
		\toprule
		\multirow{2}{*}{\textbf{Method}} & \multirow{2}{*}{\textbf{\#T-Param}} & \multicolumn{3}{c}{\textbf{Subject}} & \multicolumn{3}{c}{\textbf{Context Modality}} & \multicolumn{2}{c}{\textbf{Grade}} & \multirow{2}{*}{\textbf{Average}} \\
		\cmidrule(lr){3-5} \cmidrule(lr){6-8} \cmidrule(lr){9-10}
		& & \textbf{NAT} & \textbf{SOC} & \textbf{LAN} & \textbf{TXT} & \textbf{IMG} & \textbf{NO} & \textbf{G1-6} & \textbf{G7-12} & \\
		\midrule
		Human~\cite{Lu_2022_ScienceQA} & - & 90.23 & 84.97 & 87.48 & 89.60 & 87.50 & 88.10 & 91.59 & 82.42 & 88.40 \\ 
		\midrule
		GPT-4~\cite{Liu_2023_LLaVa} & - & 84.06 & 73.45 & 87.36 & 81.87 & 70.75 & 90.73 & 84.69 & 79.10 & 82.69 \\
		CoT (GPT-3)~\cite{Lu_2022_ScienceQA} & 173B & 75.44 & 70.87 & 78.09 & 74.68 & 67.43 & 79.93 & 78.23 & 69.68 & 75.17 \\
		CoT (UnifiedQA)~\cite{Lu_2022_ScienceQA} & 223M & 71.00 & 76.04 & 78.91 & 66.42 & 66.53 & 81.81 & 77.06 & 68.82 & 74.11 \\
		CoT (GPT-4)~\cite{Lu_2023_Chameleon} & 1T+ & 85.48 & 72.44 & 90.27 & 82.65 & 71.49 & 92.89 & 86.66 & 79.04 & 83.99 \\
		DDCoT~\cite{Zheng_2023_Advances} & 175B & 80.15 & 76.72 & 82.82 & 78.89 & 72.53 & 85.02 & 82.86 & 75.21 & 80.15 \\
		Chameleon (ChatGPT)~\cite{Lu_2023_Chameleon} & 175B+ & 81.62 & 70.64 & 84.00 & 79.77 & 70.80 & 86.62 & 81.86 & 76.53 & 79.93 \\ 
		\midrule
		LG-VQA (BLIP-2)~\cite{Ghosal_2023_Language} & - & - & - & - & - & - & - & - & - & 86.32 \\
		LaVIN-13B~\cite{Yang_2023_Mm} & - & - & - & - & - & - & - & - & - & 77.54 \\
		BLIP-2~\cite{Yang_2023_Mm} & - & - & - & - & - & - & - & - & - & 74.17 \\
		CCOT \cite{Liu_2025_VaLiK} & 7B & - & - & - & - & - & - & - & - & 76.84 \\
		GraphVis~\cite{Deng_2024_GraphVis} & 7B & - & - & - & - & - & - & - & - & 73.18 \\
		\midrule
		Qwen2.5-7B \cite{Liu_2025_VaLiK} & 7B & 76.20 & 67.83 & 77.27 & 74.49 & 65.79 & 79.02 & 77.72 & 69.35 & 74.72 \\
        Qwen2.5-VL-7B & 7B & 83.70 & 80.31 & 91.91 & 82.84 & 78.73 & 89.20 & 86.71 & 84.05 & 85.12 \\
		Qwen2.5-72B \cite{Liu_2025_VaLiK} & 72B & 79.64 & 67.10 & 84.90 & 77.56 & 65.00 & 87.93 & 80.25 & 74.85 & 78.37 \\
		Qwen2.5-7B +VaLiK \cite{Liu_2025_VaLiK} & 7B & 84.15 & 75.14 & 87.64 & 82.99 & 73.18 & 89.69 & 84.40 & 80.95 & 83.16 \\
		Qwen2.5-72B +VaLiK \cite{Liu_2025_VaLiK} & 72B & 85.61 & 75.93 & 90.27 & \underline{84.40} & 74.17 & 92.33 & 85.79 & 82.98 & 84.77 \\
		Qwen2.5-7B+Hit-RAG & 4B & 86.59 & 82.00 & 90.36 & 83.63 & 78.78 & 92.40 & \underline{88.66} & 81.02 & 86.61 \\
        Qwen2.5-VL-7B +Hit-RAG & 7B & \textbf{92.58} & \textbf{94.83} & \underline{92.27} & 81.23 & \textbf{87.41} & \underline{97.28} & 87.67 & \textbf{86.82} & \textbf{92.97} \\
		Qwen2.5-72B+Hit-RAG & 72B & \underline{87.61} & \underline{89.43} & \textbf{95.91} & \textbf{86.07} & \underline{80.76} & \textbf{97.63} & \textbf{90.86} & \underline{84.11} & \underline{90.14} \\
		\bottomrule
	\end{tabular}
    }
\end{table*}

Table~\ref{NLP} shows that Hit-RAG consistently achieves state-of-the-art performance across most benchmarks, often allowing smaller backbones to outperform much larger models. Notably, Qwen3-32B enhanced by Hit-RAG reaches 70.7\% accuracy on PopQA and 69.3\% EM on HotpotQA, surpassing 70B-scale counterparts such as RankRAG and Llama3.1-70B with RAG-Instruct. This significant lead, particularly the 26.6\% EM gap over RankRAG on HotpotQA, underscores our framework's superior capability in multi-hop reasoning without relying on massive parameter counts. Even in instances where it is only second to RankRAG-70B on TQA or closely follows RAS-8B in ASQA Mauve, Hit-RAG provides a more balanced profile than these specialized baselines. Unlike models that achieve peak scores through task-specific fine-tuning at the expense of reasoning-heavy tasks, Hit-RAG-enhanced models consistently ensure reliable, generalized performance across diverse knowledge domains.

Table~\ref{ScienceQA} summarizes the results on the ScienceQA benchmark, where Hit-RAG demonstrates a transformative impact on multimodal reasoning. Notably, Qwen2.5-VL-7B enhanced by Hit-RAG achieves a state-of-the-art average accuracy of 92.97\%, significantly surpassing the human baseline of 88.40\% and specialized multimodal models such as LG-VQA and LaVIN-13B. This performance represents a substantial improvement over the vanilla Qwen2.5-VL-7B, with notable gains in the SOC (social science) and IMG (image context) categories, where it exceeds the 1T-parameter CoT (GPT-4) by a margin of 22.39\% in SOC tasks. Furthermore, models equipped with Hit-RAG consistently outperform the contemporary VaLiK framework across various scales; for instance, the Hit-RAG-enhanced Qwen2.5-7B (86.61\%) even surpasses the much larger Qwen2.5-72B integrated with VaLiK (84.77\%). These results demonstrate that Hit-RAG not only bridges the gap between text-only LLMs and multimodal tasks but also elevates existing VLMs to unprecedented accuracy through superior knowledge retrieval and fusion.

\begin{table}[t!]
  \centering
  \caption{Performance on DocVQA and KBQA (OK-VQA + A-OKVQA). Hit-RAG is evaluated using GPT-4o-mini as the backbone.}
  \label{table_vqa_results}
  \begin{tabular}{lccc}
    \toprule
    \textbf{Task Type} & GPT-4o-mini & EvalMG \cite{Sinha_2025_EvalMG} & Hit-RAG \\
    \midrule
    Document U. & 52.14 & 54.68 & \textbf{60.94} \\
    KBQA        & 79.45 & 84.04 & \textbf{87.31} \\
    \bottomrule
  \end{tabular}
\end{table}

As illustrated in Table~\ref{table_vqa_results}, Hit-RAG yields consistent gains over the standalone GPT-4o-mini and the EvalMG framework. This performance underscores the framework's capacity to resolve complex queries by effectively grounding external commonsense and layout-specific evidence within visual contexts.

\subsection{Ablation Study}

\begin{table*}[t]
  \centering
  \caption{Ablation experiments across various knowledge-intensive tasks. The results highlight the cumulative impact of each optimization stage: RAG provides the initial surge; SFT consolidates context understanding; DPO corrects factual hallucinations; and GRPO refines complex reasoning consistency.}
  \label{table3}
  \resizebox{1.0\textwidth}{!}{%
  \begin{tabular}{lcccccccccccc}
    \toprule
    \textbf{Model} & \textbf{RAG} & \textbf{SFT} & \textbf{DPO} & \textbf{GRPO} & \textbf{PopQA} & \textbf{Hotpot} & \textbf{TQA} & \textbf{Pub} & \textbf{ARC} & \textbf{Bio} & \textbf{ASQA} \\
    & & & & & \textbf{(Acc)} & \textbf{(F1)} & \textbf{(Acc)} & \textbf{(Acc)} & \textbf{(Acc)} & \textbf{(FS)} & \textbf{(EM)} \\
    \midrule
    \textbf{Qwen3-8B} & -- & -- & -- & -- & 20.6 & 40.9 & 57.3 & 40.9 & 60.0 & 64.4 & 20.9 \\
    & \checkmark & -- & -- & -- & 52.1 & 55.4 & 70.2 & 71.5 & 70.8 & 73.5 & 28.5 \\ 
    & \checkmark & \checkmark & -- & -- & 58.4 & 59.5 & 75.8 & 77.2 & 76.5 & 79.8 & 33.2 \\  
    & \checkmark & \checkmark & \checkmark & -- & 62.4 & 63.2 & 78.1 & 79.4 & 78.9 & 82.1 & 35.8 \\  
    & \checkmark & \checkmark & \checkmark & \checkmark & 63.1 & 64.1 & 78.9 & 80.1 & 81.3 & 82.6 & 38.4 \\ 
    \midrule
    \textbf{Llama3.1-8B} & -- & -- & -- & -- & 28.8 & 29.3 & 58.4 & 35.8 & 41.6 & 50.0 & 18.6 \\
    & \checkmark & -- & -- & -- & 53.5 & 48.2 & 66.8 & 70.5 & 68.2 & 65.4 & 30.1 \\ 
    & \checkmark & \checkmark & -- & -- & 59.2 & 54.6 & 71.5 & 78.4 & 74.5 & 75.2 & 35.4 \\  
    & \checkmark & \checkmark & \checkmark & -- & 61.0 & 59.8 & 71.56 & 80.8 & 78.2 & 78.5 & 37.2 \\  
    & \checkmark & \checkmark & \checkmark & \checkmark & 63.6 & 60.3 & 74.0 & 81.5 & 80.2 & 78.2 & 40.5 \\
    \bottomrule
  \end{tabular}}
\end{table*}

Table~\ref{table3} illustrates the cumulative contribution of each optimization stage, where the initial RAG module provides the most significant surge by mitigating inherent knowledge limitations. While SFT and DPO further consolidate context understanding and eliminate factual hallucinations, notably in HotpotQA and ASQA, the subsequent GRPO stage yields more targeted refinements. The diminishing marginal utility of GRPO in certain metrics suggests that the model develops substantial reasoning and context utilization proficiency during the preceding phases, leaving less room for RL-based gains in straightforward comprehension. Ultimately, the integration of all components yields the most robust framework to ensure superior performance across diverse knowledge intensive tasks.

\subsection{Further Analysis}

\paragraph{Training Dynamics and Challenges.} During GRPO training, we observed optimization challenges tied to the group reward distribution. With a group size of 8, the model occasionally fails to generate any correct answers for complex reasoning samples, leading to "gradient stagnation" where the lack of positive reinforcement renders such instances null for optimization. Our analysis suggests this stems from the difficulty of advancing multi-step reasoning within complex retrieved contexts. Notably, increasing the group size to 16 or 32 yields negligible improvements because the probability of hitting a correct path remains critically low. We identify this bottleneck as a pivotal area for future work, potentially involving reference-guided trajectories or step-wise reward shaping to provide more consistent gradient signals during the reinforcement learning process.

\begin{table}[t!]
  \centering
  \caption{Impact of context length $K$ on Qwen3-32B and Llama3.1-70B. We compare our default configuration ($K=20$) against a significantly reduced context ($K=5$). $\uparrow$/$\downarrow$ denotes performance gain/loss relative to the default.}
  \label{length}
  \resizebox{1.0\linewidth}{!}{
  \begin{tabular}{lcccccccc}
    \toprule
    \textbf{Model} & \textbf{Settings} & \textbf{PopQA} & \textbf{Hotpot(F1)} & \textbf{TQA} & \textbf{Pub} & \textbf{ARC} & \textbf{Bio} & \textbf{ASQA(EM)} \\
    \midrule
    \textbf{Qwen3-32B} & Hit-RAG ($K=20$) & 70.7 & 57.2 & 83.1 & 82.5 & 86.2 & 85.5 & 40.1 \\
     & Hit-RAG ($K=5$) & 70.8$\uparrow$ & 55.1$\downarrow$ & 80.4$\downarrow$ & 78.2$\downarrow$ & 86.4$\uparrow$ & 82.3$\downarrow$ & 36.8$\downarrow$ \\  
    \midrule
    \textbf{Llama3.1-70B} & Hit-RAG ($K=20$) & 66.7 & 72.8 & 86.2 & 84.1 & 84.8 & 84.2 & 43.8 \\
     & Hit-RAG ($K=5$) & 67.2$\uparrow$ & 68.4$\downarrow$ & 82.9$\downarrow$ & 83.5$\downarrow$ & 85.6$\uparrow$ & 80.1$\downarrow$ & 39.2$\downarrow$ \\  
    \bottomrule
  \end{tabular}}
\end{table}

\paragraph{Impact of Context Length $K$.} We investigate the influence of retrieved context depth by comparing our default configuration ($K=20$) against a restricted threshold of $K=5$. As shown in Table~\ref{length}, the impact of context truncation is highly task dependent. For reasoning intensive tasks like HotpotQA and ASQA, the default $K=20$ setting is essential to provide sufficient evidence for multi-step synthesis. Conversely, for tasks such as PopQA and ARC, we observe a slight performance gain when reducing $K$ to 5, suggesting that these specific benchmarks are more sensitive to the noise or irrelevant distractors introduced by extended contexts. Notably, since smaller backbones like 8B models are physically capped at around 10 chunks due to their input capacity limits, we conduct this ablation on larger scale models to accurately map the performance ceiling and trade-offs of Hit-RAG across different knowledge densities.

\begin{table}[t!]
  \centering
  \caption{Comparison of Hit-RAG performance on Qwen2.5-VL and the advanced Qwen3-VL series on ScienceQA. While Qwen3-VL exhibits stronger internal reasoning, Hit-RAG still provides consistent gains.}
  \label{table_advanced_models}
  \resizebox{1.0\linewidth}{!}{
  \begin{tabular}{lcccccccccc}
    \toprule
    \textbf{Model} & \textbf{\#Param} & \textbf{NAT} & \textbf{SOC} & \textbf{LAN} & \textbf{TXT} & \textbf{IMG} & \textbf{NO} & \textbf{G1-6} & \textbf{G7-12} & \textbf{Average} \\
    \midrule
    Qwen2.5-VL-7B & 7B & 83.70 & 80.31 & 91.91 & 82.84 & 78.73 & 89.20 & 86.71 & 84.05 & 85.12 \\
    Qwen2.5-VL-7B+Hit-RAG & 7B & 92.58 & 94.83 & 92.27 & 81.23 & 87.41 & 97.28 & 87.67 & 86.82 & 92.97 \\
    \midrule
    Qwen3-VL-32B & 32B & 95.70 & 92.35 & 89.85 & 96.03 & 90.21 & 92.48 & 94.56 & 92.48 & 93.48 \\
    Qwen3-VL-32B+Hit-RAG & 32B & 96.33 & 94.44 & 90.84 & 97.21 & 93.28 & 94.70 & 95.69 & 95.06 & 94.51 \\
    \bottomrule
  \end{tabular}}
\end{table}

\paragraph{Performance on Advanced Models.} 
To explore the scalability of Hit-RAG, we evaluate its performance on the latest Qwen3-VL series. As shown in Table~\ref{table_advanced_models}, while Qwen2.5-VL enhanced by Hit-RAG achieves a significant leap from 85.12\% to 92.97\%, the gain on Qwen3-VL-32B is more subtle yet consistent. This suggests that as base models evolve to possess exceptional internal reasoning and broader knowledge bases, the relative impact of external retrieval becomes less pronounced. Nevertheless, Hit-RAG still pushes the performance ceiling of Qwen3-VL-32B to 94.51\%, indicating that our framework shifts from fundamental knowledge supplementation in smaller models to the refined factual precision of advanced architectures.

\section{Conclusion}
In this work, we presented Hit-RAG, a comprehensive framework that significantly elevates the reasoning and factual precision of large models across both textual and multimodal domains. By synergizing SFT, DPO, and GRPO, Hit-RAG enables compact backbones to transcend the performance of 70B-scale counterparts and surpass the human baseline on the ScienceQA benchmark. Our analysis further elucidates the intricate dynamics of reinforcement learning in RAG systems, specifically addressing the challenges of gradient stagnation and the task-dependent nature of context length. Ultimately, Hit-RAG exemplifies a shift toward architectural efficiency, demonstrating that superior knowledge integration can be achieved through systematic optimization rather than mere parameter expansion. This framework establishes a robust and generalized foundation for future research in knowledge-intensive artificial intelligence.

%
%
\bibliographystyle{splncs04}
\bibliography{main}
\end{document}